\documentclass[cameraready]{Interspeech}
\usepackage{booktabs}
\usepackage{multirow}
\usepackage{xcolor}
\usepackage{subcaption}
\usepackage{colortbl}
\usepackage{amsmath}

\usepackage{siunitx}
\usepackage{graphicx}
\usepackage{numprint}

\npdecimalsign{.}
\nprounddigits{2}
\npstyleenglish

\title{Linguistic Bias Mitigation for Spoofing Detection via \\ Gradient Reversal and A Variational Information Bottleneck}

\author[affiliation={1}]{Anh-Tuan}{Dao}
\author[affiliation={1}]{Driss}{Matrouf}
\author[affiliation={1}]{Mickael}{Rouvier}
\author[affiliation={2}]{Nicholas}{Evans}

\address{
    $^1$ Laboratoire Informatique d’Avignon, Avignon Universite, France \\
    $^2$ EURECOM, Sophia Antipolis, France 
}
\email{\{anh-tuan.dao, driss.matrouf, mickael.rouvier\}@univ-avignon.fr, evans@eurecom.fr}


\usepackage{comment}

\begin{document}

\maketitle

\begin{abstract}

Rapid advancements in generative speech technology have compromised the reliability of voice biometrics. While current spoofing detectors excel when assessed under in-domain conditions, generalisation to out-of-domain settings is often poor. We show that this can be due to linguistic bias. A reliance on linguistic cues observed in training data can then compromise robustness to cross-data. We propose a linguistic-invariant spoofing detection framework utilizing teacher–student adversarial learning. The linguistic-aware teacher model, pre-trained on linguistic content of an external dataset, guides the student detector via gradient reversal to minimize the linguistic information. To prevent the inadvertent removal of non-linguistic cues, we incorporate a Variational Information Bottleneck to enable suppression of principal cues. Across nine DF Arena datasets, our method achieves up to a 36.2\% relative reduction in the EER compare to the baseline.

\end{abstract}

\section{Introduction}
Advances in text-to-speech (TTS) and voice conversion have led to realistic speech deepfakes, posing security risks to the reliability of voice biometrics. To mitigate these threats, spoofing detection systems are deployed to protect voice biometrics. In recent years, the ASVspoof and Audio Deepfake Detection (ADD) initiatives have driven substantial advances in spoofing detection through the collection and distribution of datasets and through benchmarking evaluations~\cite{ASVspoof24,ADD2022}.
Recent studies have shown that self-supervised learning (SSL) models significantly improve spoofing detection performance by leveraging large-scale pre-training using raw speech~\cite{AASIST,conformer,rohdin24_but_asvspoof,MHFA_Spoof,SLS,Mamba}. Fine-tuned SSL representations have emerged as the state-of-the-art paradigm, leveraging high-dimensional acoustic features to achieve superior generalization compared to conventional end-to-end supervised architectures~\cite{AASIST1,RawNet2,dao24_asvspoof}. Despite these advances, generalization remains a central challenge: models that perform well on a given benchmark often fall short when applied to unseen datasets or different evaluation conditions.

Growing evidence suggests that spoofing detection instability stems from shortcut learning whereby biased models are able to exploit spurious correlations rather than genuine spoofing artifacts. For example, M\"uller et al.~\cite{Silence2021} identified a non-speech-related shortcut in ASVspoof~2019, where bona fide utterances consistently included leading silence while spoofed samples did not. Models that rely on such spurious artifacts then achieve high in-domain accuracy but suffer from poor out-of-domain generalization. The ASVspoof~5 dataset was introduced to mitigate these limitations, providing a substantially larger corpus with diverse attack types and codecs~\cite{ASVspoof24}. While spoofing detection systems trained and tested using the ASVspoof~5 database achieve strong detection performance, recent results show that the augmentation of ASVspoof~5 training data with data sourced from additional spoofing corpora often leads to performance degradation~\cite{wang2026asvspoof5evaluationspoofing}. This unexpected observation suggests the presence of overlooked dataset-specific biases.

In this work, we identify linguistic bias in the ASVspoof~5 dataset, manifested as differences in spoken content between bona fide and spoofed utterances. 
When such bias exists, spoofing detection models may associate specific sentence patterns with spoofed or bona fide labels, allowing them to make predictions based on what is being said rather than how the speech is generated. This is harmful for generalization, as linguistic distributions can vary across datasets.
To mitigate this issue, we propose a framework that suppresses linguistic information during spoofing detection training. Our approach is based on a teacher–student framework with adversarial learning. Since text transcripts are unavailable for the ASVspoof~5 database, we first train a teacher model using a Common Voice subset for phrase linguistic content classification so that generated embeddings encode linguistic information. In both the teacher and the student, SSL representations are
aggregated with multi-head factorized attentive pooling (MHFA)~\cite{MHFA_ASV}. The student spoofing detection model is then guided by the teacher through a Gradient Reversal Layer (GRL)~\cite{pmlr-2015-GRL,dao26_assessing_spk,dao2026enhancing}, encouraging the student to learn representations that are invariant to linguistic content while remaining discriminative for spoofing detection.

While teacher embeddings encode linguistic content, they are entangled with latent features.
Adversarial suppression can be overly aggressive: without additional constraints, the
student may remove not only linguistic information but also other correlated cues that
are beneficial to spoofing detection. To mitigate this issue, we further introduce a
Variational Information Bottleneck (VIB)~\cite{alemi2016deepVIB} in the student model.
The VIB constrains the amount of information removed during adversarial training,
enforcing a bottleneck to suppress dominant linguistic information while preserving
spoofing-discriminative evidence. This combination facilitates controlled and principled
invariant learning.


We make the following contributions.
\begin{itemize}
    \item We identify and empirically demonstrate an unexplored linguistic shortcut in the ASVspoof~5 dataset, revealing a linguistic bias that undermines generalization.
    \item We propose a teacher–student framework with adversarial GRL training to suppress linguistic information without requiring text annotations for the test dataset.
    \item We show that the VIB can be used to control the removal of reliable spoofing cues while still suppressing linguistic cues.
    \item Extensive experiments show consistent and substantial improvements over nine evaluation datasets.
\end{itemize}

\section{Linguistic Bias Analysis}
\label{sec2}

\begin{figure}[t]

    \centering

    \includegraphics[width=0.9\linewidth]{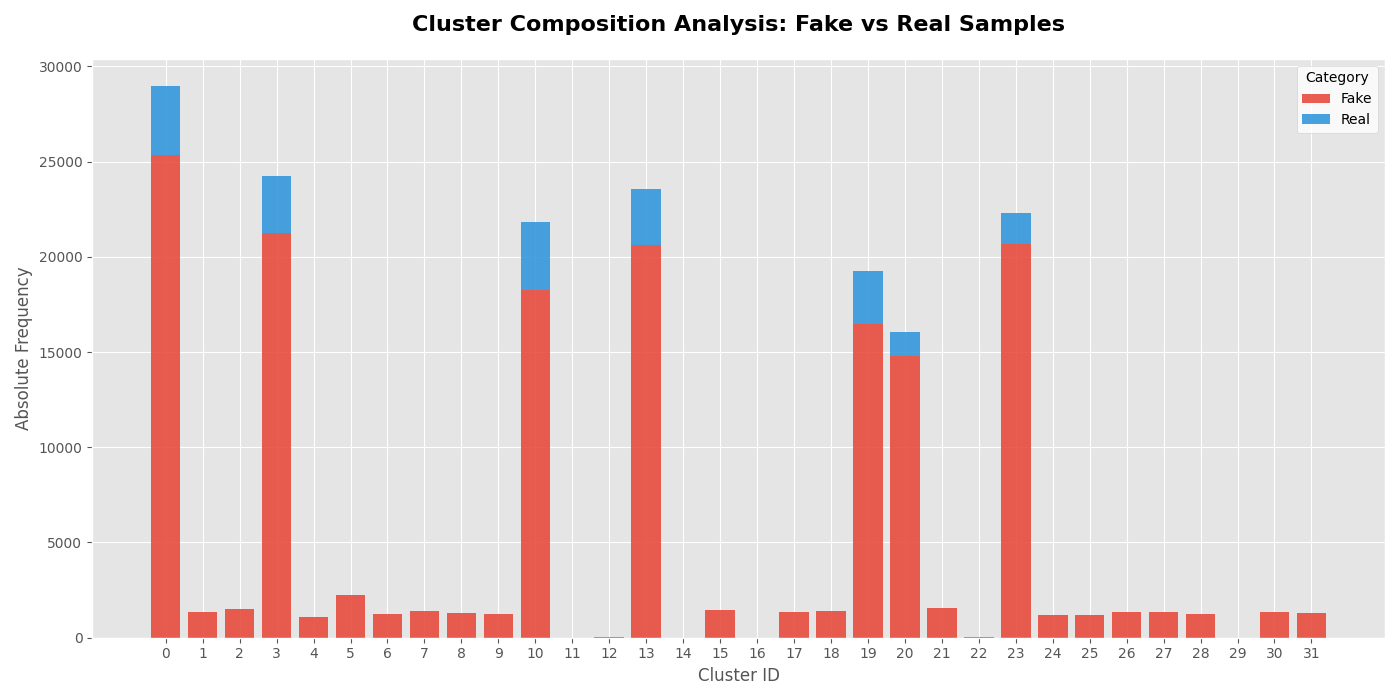}   
    \caption{Cluster composition analysis of \textbf{SONAR embeddings} for ASVspoof 5 training data.}
    \label{fig:cluster1}
\end{figure}

\begin{figure}[t]

    \centering

    \includegraphics[width=0.9\linewidth]{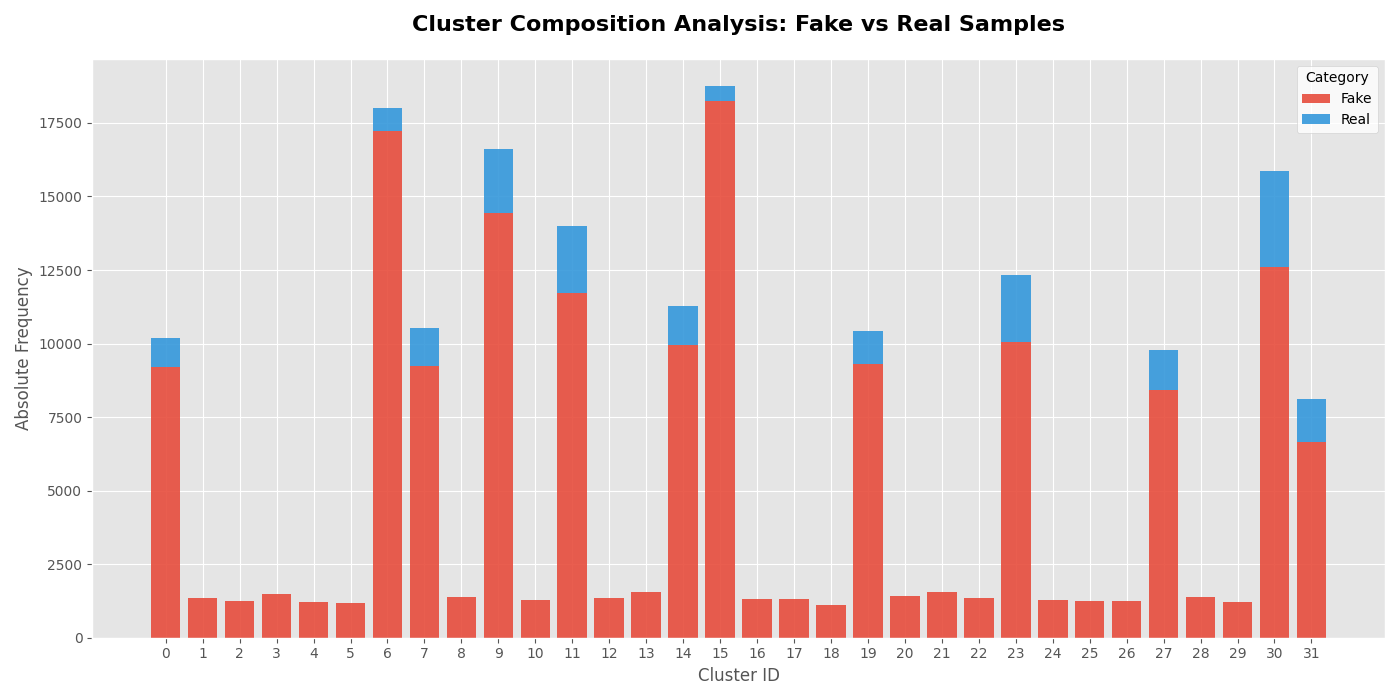}   
    \caption{Cluster composition analysis of phrase linguistic content  \textbf{teacher embeddings} for ASVspoof 5 training data.}
    \label{fig:cluster2}
\end{figure}

Spoofing detection corpora can suffer from linguistic bias when spoofed and bona fide utterances differ in spoken content. In such cases, classifiers may learn to associate specific linguistic patterns with class labels, allowing them to rely on what is being said rather than acoustic artifacts indicative of spoofing. 
We investigate the presence of linguistic bias in the ASVspoof~5 training set by analysing embeddings generated by the SONAR model~\cite{Duquenne:2023:sonar_arxiv}. 
While SONAR is optimized for semantic proximity, we utilize its embeddings as a high-level proxy for linguistic content.
To further validate our findings, we complement this approach with a custom phrase linguistic content classification model (detailed in Section \ref{Sec:31}) specifically designed to extract linguistic embeddings.
We then apply k-means clustering ($K=32$) to the extracted SONAR and phrase linguistic content embeddings, and analyse the resulting cluster distributions (Figures \ref{fig:cluster1}, \ref{fig:cluster2}). Our analysis reveals a structural imbalance of clusters: while bona fide samples are concentrated within a specific subset of clusters, a significant majority of clusters are comprised exclusively of spoofed utterances. These distributions confirm a linguistic mismatch between the bona fide and spoofed partitions of the ASVspoof~5 training data which may result in imperfect models derived from the training process.

Figure \ref{fig:learned_emb} presents t-SNE visualizations of phrase linguistic content classification teacher embeddings for a random speaker in ASVspoof~5 train. 
In ASVspoof~5, spoofed samples form distinct clusters. 
This pattern mirrors the dataset generation process, which utilized eight TTS models, resulting in eight-sample clusters.
Conversely, bona fide samples reside far from these spoofed clusters, indicating a severe text content mismatch.
This visualization exposes a critical spoken content misalignment between bona fide and spoofed audio.
Such issues risk the model relying on linguistic artifacts rather than robust spoofing cues, compromising performance across diverse datasets.

\begin{figure}[t]
\label{fig:learned_emb}
    \centering

    \includegraphics[width=0.9\linewidth]{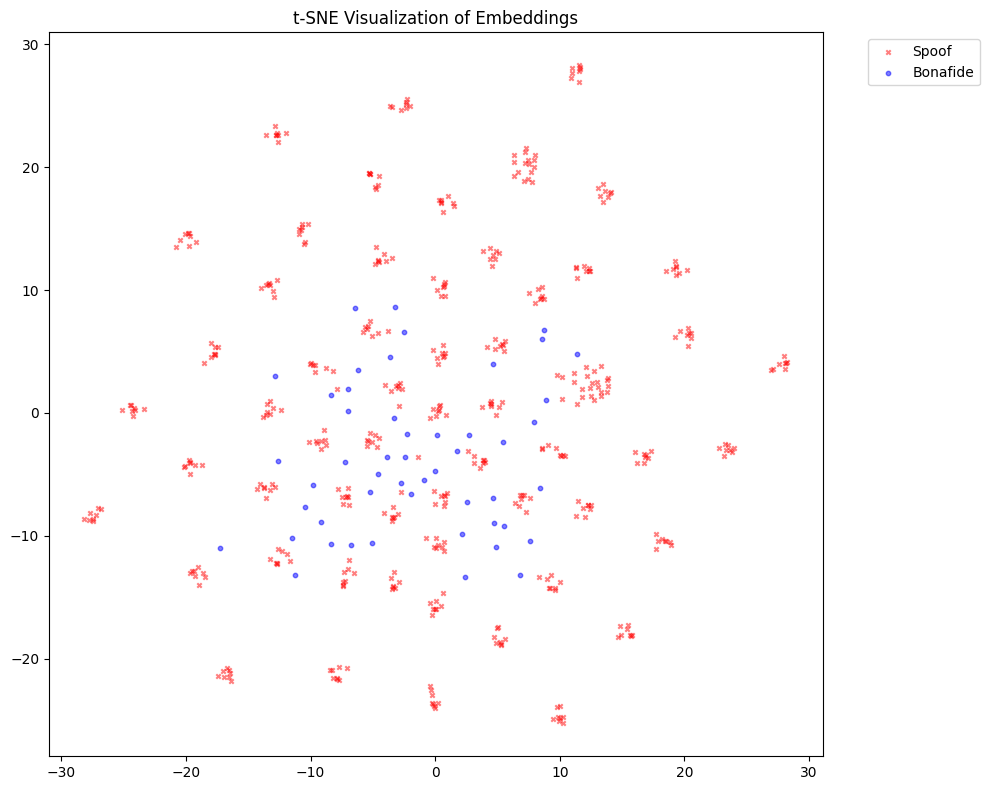}
    
    \caption{t-SNE visualization of {phrase linguistic content \textbf{teacher embeddings}} for a random speaker in ASVspoof~5 train.} 
    \label{fig:learned_emb}
\end{figure}

\section{Method for Linguistic Bias Mitigation}
We describe in the following our use of a teacher-student framework to suppress the use of linguistic information for spoofing detection. The teacher model is trained using a phrase linguistic content classification task to capture linguistic features. The student model is trained in a multi-task manner for both spoofing detection and phrase linguistic content classification, incorporating an adversarial learning mechanism to minimize the influence of linguistic information on spoofing detection.

\subsection{Phrase linguistic content classification teacher model}
\label{Sec:31}
While SONAR model effectively identifies linguistic bias (Section \ref{sec2}), its massive parameter scale and distinct architecture complicate its integration into a teacher-student framework for spoofing detection. Accordingly, we propose an architecturally compatible teacher model: an XLSR-based feature extractor coupled with a Multi-Head Factorized Attentive (MHFA) backend classifier \cite{MHFA_Spoof}.

To model linguistic information, we train a phrase linguistic content teacher model using the English subset of the Common Voice dataset, which contains speech recordings collected from a large number and variety of speakers and with phonetically diverse content. We curate a training dataset using Common Voice English transcriptions, comprising 10.3k unique phrases across 158k utterances. 
The teacher model is trained to classify input utterances into one of the distinct text phrase IDs. Once trained, the teacher model is frozen and used solely as a reference embedding extractor, providing soft labels to guide the student model.

\subsection{Spoofing detection student model}
\begin{figure}[t]
\centering
\includegraphics[width=0.8\linewidth]{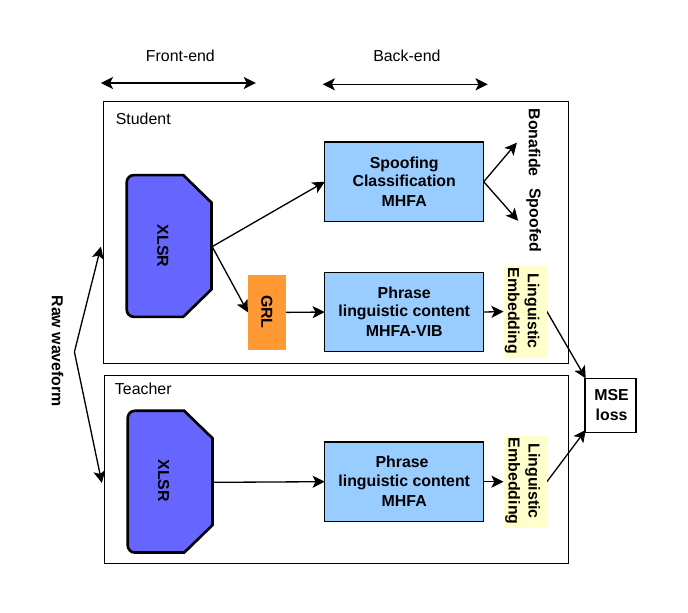}
\caption{The proposed IVLing-VIB model for invariant
linguistic within a teacher-student framework.}

\label{fig:idfe_arch}
\end{figure}

The student model is jointly optimized for spoofing detection and phrase linguistic content classification tasks, with an adversarial component to disentangle linguistic information from the learned representations.
As depicted in Figure~\ref{fig:idfe_arch}, the student model comprises the following key components:
\begin{itemize}
    \item \textbf{Feature extractor:} A pretrained XLSR encoder~\cite{XLS-R2022} which transforms raw audio waveforms into a sequence of contextualized frame-level embeddings.
    \item \textbf{Spoofing classification head:} A MHFA network which uses the XLSR embeddings to predicts the binary label $\hat{y}_s\in \{\text{bonafide},\text{spoof}\}$.
    \item \textbf{Phrase linguistic content head:} A MHFA-VIB network dedicated to the phrase linguistic content classification task. This head is supervised using soft targets from the teacher model embeddings with optimization performed via a mean squared error loss between student and teacher embeddings.
    \item \textbf{Gradient reversal layer (GRL):} Positioned between the feature extractor and the phrase linguistic content head. During the forward pass, the GRL functions as an identity transformation. Under backpropagation, it scales the gradients by a factor of $-\lambda$ (where $\lambda$ is a hyperparameter), adversarially encouraging the feature extractor to produce representations that are invariant to linguistic cues.
\end{itemize}

\subsection{MHFA-VIB for linguistic invariance}

We introduce a VIB regularization into the phrase linguistic content branch of the student model, resulting in the MHFA-VIB architecture. The VIB constrains the information that can flow through the linguistic branch during adversarial training, ensuring that suppression focuses on the principal linguistic components, facilitating controlled and principled invariant learning. Given multi-layer XLSR representations
$\mathbf{o} \in \mathbb{R}^{D \times T \times L}$, where $L$
denotes the number of transformer layers, MHFA-VIB first learns
layer-wise importance weights for the key and value streams.
After softmax normalization, the weighted sum across layers is computed as
\begin{equation}
\mathbf{k} = \sum_{l=1}^{L} w_k^{(l)} \mathbf{o}^{(l)}, \quad
\mathbf{v} = \sum_{l=1}^{L} w_v^{(l)} \mathbf{o}^{(l)},
\end{equation}
followed by a linear compression to obtain
$\mathbf{k}, \mathbf{v} \in \mathbb{R}^{d_c \times T}$.

To impose an information bottleneck, the VIB module is applied
to the key representation $\mathbf{k}$.
Specifically, a stochastic latent variable $\mathbf{z}$ is modeled using a
Gaussian posterior:
\begin{equation}
q_\theta(\mathbf{z} \mid \mathbf{k}) =
\mathcal{N}(\boldsymbol{\mu}(\mathbf{k}),
\mathrm{diag}(\boldsymbol{\sigma}^2(\mathbf{k}))),
\end{equation}
where the mean $\boldsymbol{\mu}$ and log-variance
$\log \boldsymbol{\sigma}^2$ are predicted using neural networks.
Sampling is performed using the reparameterization trick:
\begin{equation}
\mathbf{z} = \boldsymbol{\mu} + \boldsymbol{\epsilon} \odot
\boldsymbol{\sigma}, \quad
\boldsymbol{\epsilon} \sim \mathcal{N}(\mathbf{0}, \mathbf{I}).
\end{equation}

The VIB is enforced through a
Kullback--Leibler (KL) divergence regularization term, defined as
\begin{equation}
\mathcal{L}_{\text{VIB}} =
\mathrm{KL}(q_\theta(\mathbf{z} \mid \mathbf{k}) \,\mid\mid\, r(\mathbf{z})),
\end{equation}
where $q_\theta(\mathbf{z} \mid \mathbf{k})$ denotes the variational
posterior over the latent representation and
$r(\mathbf{z}) = \mathcal{N}(\mathbf{0}, \mathbf{I})$ is a standard normal
prior.
This penalty induces a stochastic bottleneck, limiting the capacity of the latent variable $\mathbf{z}$ and preventing the model from encoding redundant features from $\mathbf{k}$.

The latent representation $\mathbf{z}$ is transformed into temporal attention weights, which guide the weighted aggregation of value representations $\mathbf{v}$. This process yields an utterance-level embedding tailored to the phrase linguistic content sub-task.

\subsection{Linguistic-invariant teacher-student training}

Given an input utterance $x$, the feature extractor produces frame-level representations $o = f(x)$.  
The spoofing classification head is denoted as $g_s(\cdot)$, while the phrase linguistic content head is denoted as $g_l(\cdot)$.  
A frozen pretrained teacher model $g_t(\cdot)$ provides reference linguistic embeddings.
The student model is optimized according to:
\begin{equation}
\min_{f, g_s, g_l} 
\mathcal{L}_s(g_s(f(x)), y_s)
+ \alpha \mathcal{L}_l\!(g_l(\mathrm{GRL}(f(x))),\, g_t(x))
+ \beta \mathcal{L}_{\mathrm{VIB}}
\label{eq:totalloss}
\end{equation}

\noindent where each component is

\begin{itemize}
    \item $\mathcal{L}_s$ is the cross-entropy loss for spoofing detection.
    \item $\mathcal{L}_l$ is the mean squared error loss between student and teacher linguistic embeddings.
    \item $\mathcal{L}_{\mathrm{VIB}}$ is the Kullback-Leibler divergence between the variational posterior and the prior in the VIB module.
    \item $\alpha$ controls the influence of linguistic invariance.
    \item $\beta$ controls the strength of the information bottleneck.
\end{itemize}







\begin{table*}[t]
\renewcommand{\arraystretch}{0.8} 
\centering
\small
\caption{Model performance (EER, \%) across datasets. ``Average'' reports the mean EER across evaluation sets. ``Pooled'' is obtained by combining all datasets and computing EER with a single global decision threshold, and is used as the main measure of cross-dataset generalization. For MHFA-based variants, the value in parentheses indicates the absolute gain (EER) compared to MHFA.  Bold numbers indicate the best performance, while underlined numbers denote the second best.}
\label{tab:eer}
\begin{tabular}{lcccccc}
\toprule
 & \multicolumn{6}{c}{\textbf{Model (EER \%)}} \\
\cmidrule(lr){2-7}
\textbf{Dataset} & \textbf{AASIST} & \textbf{Conformer} & \textbf{MHFA} & \textbf{MHFA-VIB} & \textbf{MHFA-IVLing} & \textbf{MHFA-IVLing-VIB} \\
\midrule
ITW        & 7.03 & 5.68 & 4.30 & 3.95 {\color{gray}(+0.35)} & \underline{1.93} {\color{gray}(+2.37)} & \textbf{1.88} {\color{gray}(+2.42)} \\
ASV~19 eval &10.79 &10.80 & 9.48 & 7.24 {\color{gray}(+2.24)} & \underline{5.04} {\color{gray}(+4.44)} & \textbf{4.07} {\color{gray}(+5.41)} \\
ASV~21 LA    &11.99 &10.93 &11.55 & 9.41 {\color{gray}(+2.14)} & \underline{5.87} {\color{gray}(+5.68)} & \textbf{5.58} {\color{gray}(+5.97)} \\
ASV~21 DF    & 5.29 & 5.54 & 4.83 & 5.49 {\color{gray}(-0.66)} & \underline{3.67} {\color{gray}(+1.16)} & \textbf{3.09} {\color{gray}(+1.74)} \\
FoR        & 5.65 &10.60 &11.52 & 6.14 {\color{gray}(+5.38)} & \underline{4.68} {\color{gray}(+6.84)} & \textbf{3.35} {\color{gray}(+8.17)} \\
CodecFake  &38.67 &30.30 &30.33 &24.69 {\color{gray}(+5.64)} & \underline{20.80} {\color{gray}(+9.53)} & \textbf{20.28} {\color{gray}(+10.05)} \\
DFADD      &10.03 & 5.82 & 2.11 & 2.80 {\color{gray}(-0.69)} & \underline{1.72} {\color{gray}(+0.39)} & \textbf{0.79} {\color{gray}(+1.32)} \\
LibriSeVox &23.18 &22.83 & 7.82 & \underline{2.99} {\color{gray}(+4.83)} & 3.04 {\color{gray}(+4.78)} & \textbf{2.15} {\color{gray}(+5.67)} \\
SONAR      &19.12 &22.60 &24.37 & \textbf{9.37} {\color{gray}(+15.00)} &19.56 {\color{gray}(+4.81)} & \underline{15.25} {\color{gray}(+9.12)} \\
\midrule
\textbf{Average}    &14.64 &13.90 &11.81 & 8.01 {\color{gray}(+3.80)} & \underline{7.37} {\color{gray}(+4.44)} & \textbf{6.27} {\color{gray}(+5.54)} \\
\textbf{Pool}       &19.98 &15.58 &13.67 &11.83 {\color{gray}(+1.84)} & \underline{9.56} {\color{gray}(+4.11)} & \textbf{8.72} {\color{gray}(+4.95)} \\
\bottomrule
\end{tabular}
\end{table*}

\section{Experimental Setup}

\subsection{Phrase linguistic content classification dataset}
We train a phrase linguistic content classification model using the Common Voice dataset~\cite{nathan2025unified}. Passphrases are grouped according to their transcriptions, and only those occurring at least five times and fewer than twenty times are retained. This filtering results in approximately 10,300 unique passphrases for training. In total, the dataset comprises approximately 158,000 utterances, enabling the training of a classification head and the extraction of embeddings that capture lexical content.

\subsection{Spoofing detection datasets}
We train all models using the ASVspoof~5 training set, which comprises approximately 180,000 utterances collected from 400 speakers. To avoid evaluation bias, all experiments are conducted exclusively on out-of-domain datasets. We follow the Speech DF Arena protocols~\cite{dfarena}, selecting only English-language datasets. This results in nine evaluation datasets: In-the-Wild (ITW)~\cite{Wild2022}, ASVspoof~2019~\cite{ASVspoof19}, ASVspoof~2021 LA and DF~\cite{ASVspoof21}, Fake-or-Real (FoR)~\cite{FoR2019}, CodecFake~\cite{codecfake}, DFADD~\cite{dfadd}, LibriSe-Vox~\cite{librisevox}, and SONAR~\cite{SONAR}.
Performance is assessed using the equal error rate (EER). 



\subsection{Implementation details}\label{sec:implem_details}
During training, utterances are randomly cropped into 4-second segments, while full-length utterances are used for evaluation. Data augmentation strategies are applied during training using the MUSAN corpus and a real room impulse response (RIR) database~\cite{MUSAN,Reverb2017}. Model optimization is performed using Adam optimizer~\cite{adam} with a learning rate of $10^{-6}$ to minimize the total loss in Equation~\ref{eq:totalloss}. Training is conducted for 30 epochs with a batch size of 32 using NVIDIA A100 GPUs. The weighting coefficients $\alpha$ and $\beta$ in Equation~\ref{eq:totalloss} are both set to 0.1. In addition to three baseline models also all based on XLSR (i.e., AASIST~\cite{AASIST}, Conformer~\cite{conformer}, and MHFA~\cite{MHFA_Spoof}), we include a variant that integrates the VIB regularization into the MHFA classifier, referred to as MHFA-VIB. All baseline models are compared against the proposed linguistic-invariant framework with and without the VIB module (i.e., IVLing and IVLing-VIB).

\section{Results}

\subsection{Baseline spoofing detection performance}

The variation in performance for baseline models AASIST, Conformer, and MHFA (Table~\ref{tab:eer}) underscores the generalization challenge in spoofing detection.
While EERs are between approximately 5 and 7\% EERs for the ASVspoof~2021 DF database, with only few exceptions, performance degrades substantially for almost all other datasets.
With the lowest pooled EER of 14\% MHFA is the better generalising of the three baselines.
With only two exceptions, integration of the VIB module (MHFA-VIB) yields consistent performance gains, most notably for SONAR for which the pooled EER falls from 24\% to 9\%, resulting in
a 13\% relative improvement in pooled EER.
These results confirm that the VIB objective produces generalizable representations by filtering task-irrelevant cues.

\subsection{Performance of linguistic-invariant models}
The IVLing model, which learns linguistic-invariant representations within a teacher–student framework, 
substantially outperforms both MHFA and MHFA-VIB models. Performance improves across all datasets, without exception. The pooled EER falls to 10\%, a relative improvement of approximately 30\% over the MHFA baseline and 19\% over the MHFA-VIB model. 
The integration of VIB into the linguistic-invariant framework (IVLing-VIB) further enhances performance. IVLing-VIB achieves the best overall results, obtaining the lowest EER for the majority of datasets (notably 1.88\% for ITW, 4.07\% for ASV 19, 0.79\% for DFADD, with substantial improvements for other datasets) and the lowest pooled EER of 9\%, an improvement of 36\% relative to the MHFA baseline and 9\% compared to the IVLing model.

\begin{table}[t]
\renewcommand{\arraystretch}{0.8} 
\centering
\small
\caption{Comparison between MHFA-IVLing-VIB and top-4 submissions to Track~1 (open condition) of the ASVspoof~5 challenge. EER (\%). Bold indicates best performance.}
\label{tab:evaluation_results}

\resizebox{\columnwidth}{!}{
\begin{tabular}{lccccc}
\toprule
 & \multicolumn{5}{c}{\textbf{Model (EER \%)}} \\
\cmidrule(lr){2-6}
\textbf{Dataset} & \textbf{T43} & \textbf{T27} & \textbf{T36} & \textbf{T23} & \textbf{MHFA-IVLing-VIB} \\
\midrule
ASV~5 eval & 4.33 & \textbf{3.30} & 3.37 & 4.23 & 5.26 \\
ASV~19 LA  & 26.63 & 17.33 & 16.27 & 16.73 & \textbf{4.07} \\
ASV~21 LA  & 25.57 & 18.70 & 15.73 & 13.13 & \textbf{5.58} \\
ASV~21 DF  & 14.20 & 10.63 & 11.57 & 14.87 & \textbf{3.09} \\
ITW        & 6.85  & 13.37 & 14.71 & 10.20 & \textbf{1.88} \\
\midrule
\textbf{Average} & 15.51 & 12.66 & 12.33 & 11.83 & \textbf{3.97} \\
\bottomrule
\end{tabular}}
\end{table}

\subsection{Benchmarking against top challenge submissions}
In Table~\ref{tab:evaluation_results} we present a comparison of results for the IVLing-VIB model to that of the top-4 ASVspoof~5 Challenge submissions reported in~\cite{wang2026asvspoof5evaluationspoofing}.
Results show a consistent, substantial gap in performance for ASVspoof~5 and other databases.
In contrast, a comparatively modest increase in the EER for the IVLing-VIB model on the ASVspoof~5 database is offset by a substantial gain in the EER for all other datasets.

\section{Conclusions}
In this paper, we address a critical yet unexplored limitation of spoofing detection systems: linguistic bias. We demonstrate that, in ASVspoof~5, mismatches in spoken content between spoofed and bona fide utterances enable models to exploit linguistic cues, undermining generalization. 
To address this, we propose IVLing-VIB, a teacher–student adversarial framework designed to mitigate these biases by enforcing linguistic invariance.
Using a linguistic-aware teacher and a gradient reversal layer, our approach suppresses deceptive linguistic information without requiring text annotations. Crucially, the integration of a Variational Information Bottleneck regulates this process, managing the inherent trade-off between removing biased linguistic features and preserving essential acoustic cues for spoofing detection.
The resulting IVLing-VIB model consistently outperforms baselines across nine evaluation datasets, achieving the lowest pooled EER with a 36.2\% relative gain.

\section{Acknowledgements}

This work was performed using HPC resources from GENCI-IDRIS. This work was financially supported by ANR BRUEL (ANR-22-CE39-0009).

\section{Generative AI Use Disclosure}
During the preparation of this work, the authors used generative
AI models (specifically Google Gemini and ChatGPT) to polish the language and correct grammatical errors in this manuscript. The authors rigorously reviewed and edited all text and take full responsibility for the publication content.

\bibliographystyle{IEEEtran}
\bibliography{mybib}

\end{document}